\documentclass{article}
\usepackage[preprint]{spconf}

\usepackage{calc}
\usepackage{xspace}
\usepackage{amsfonts}
\usepackage{amsmath}
\usepackage{amssymb}

\usepackage{csvsimple}
\usepackage{hyperref}
\usepackage[%
square,%
numbers%
]{natbib}
\usepackage{tabularx}
\usepackage{graphicx}
\usepackage{textcomp}

\usepackage{caption}
\usepackage{cleveref}

\makeatletter
\newcommand\footnoteref[1]{\protected@xdef\@thefnmark{\ref{#1}}\@footnotemark}
\makeatother

\newcommand\MyGlobalColorMap{Greys}

\title{Image-Based Correction of Continuous and Discontinuous Non-Planar Axial
    Distortion in Serial Section Microscopy}

\name{Philipp Hanslovsky, John Bogovic, and Stephan Saalfeld}
\address{HHMI Janelia Research Campus, 19700 Helix Dr, Ashburn, VA 20147, USA%
}

\DeclareMathOperator*{\argmin}{arg\,min}

\newcommand{\zref}{z_{\text{ref}}}
\newcommand\RangedWindow[2]{w_{#1}(#2)}
\newcommand\RRangedWindow[1]{\RangedWindow{r}{#1}}
\newcommand{\fitWindowWeight}{\RangedWindow{f}{z,\zref}}
\newcommand{\rhom}{\mathbf{S}}
\newcommand{\rhomTransformed}{\mathcal{S}}
\newcommand\zmap[1][z]{c(#1)}
\newcommand\zmult[1][z]{m(#1)}

\newcommand{\similarityEstimate}{\bar{s}}

\makeatletter
\DeclareRobustCommand\onedot{\futurelet\@let@token\@onedot}
\def\@onedot{\ifx\@let@token.\else.\null\fi\xspace}

\def\eg{\emph{e.g}\onedot} 
\def\ie{\emph{i.e}\onedot} 
\def\cf{\emph{c.f}\onedot}

\makeatother

\newcommand\DataResolution[4]{#1\,\texttimes\,#2\,\texttimes\,#3\,#4\ensuremath{^\text{3}}}
\newcommand\PhysicalResolution[3]{\DataResolution{#1}{#2}{#3}{nm}}

\newcommand\PixelResolution[3]{\DataResolution{#1}{#2}{#3}{px}}
\newcommand\UnitRange[3]{#1\,#3 to #2\,#3}
\newcommand\UnitPair[3]{#1\,#3 and #2\,#3}

\newlength{\OnePt}
\begin{document}

\maketitle

\abstract{%
    \textbf{Motivation:} Serial section microscopy is an established method for detailed anatomy reconstruction of biological specimen.
    During the last decade, high resolution electron microscopy (EM) of serial sections has become the de-facto standard for reconstruction of neural connectivity at ever increasing scales (EM connectomics).
    In serial section microscopy,
    the axial dimension of the volume is sampled by physically removing thin sections from the embedded specimen and subsequently imaging either the block-face or the section series.
    This process has limited precision leading to inhomogeneous non-planar sampling of the axial dimension of the volume which, in turn, results in distorted image volumes.  This includes that section series may be collected and imaged in unknown order.\\
    \textbf{Results:} We developed methods to identify and correct these distortions through image-based signal analysis without any additional physical apparatus or measurements.  We demonstrate the efficacy of our methods in proof of principle experiments and application to real world problems.\\
    \textbf{Availability and Implementation:} We made our work available as libraries for the ImageJ distribution Fiji and for deployment in a high performance parallel computing environment. Our sources are open and available at \href{http://github.com/saalfeldlab/section-sort}{http://github.com/saalfeldlab/section-sort}, \href{http://github.com/saalfeldlab/em-thickness-estimation}{http://github.com/ saalfeldlab/em-thickness-estimation}, and \href{http://github.com/saalfeldlab/z-spacing-spark}{http://github.com/ saalfeldlab/z-spacing-spark}.\\
    \textbf{Contact:} \href{saalfelds@janelia.hhmi.org}{saalfelds@janelia.hhmi.org}
}

\maketitle

\section{Introduction}

\begin{figure*}[]
\includegraphics[width=\linewidth]{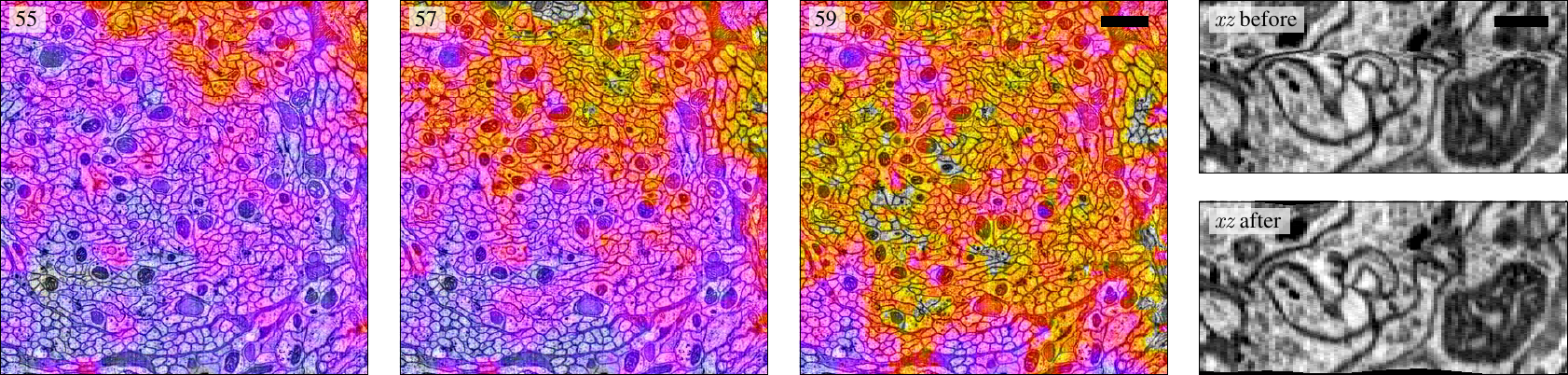}
\caption{Left: Three FIB-SEM $xy$-section scans showing \emph{Drosophila melanogaster} neural tissue overlaid with
    color-coded local $z$-spacing, serial index top left.  Color overlay was chosen arbitrarily to
    visualize the wave-like evolution of height variance.  Scale bar 1\,\textmu{}m.  Right:
    Magnified crop of an $xz$-cross-section of the original (top) and corrected (bottom) series,
    $z$-compression by the ``wave'' is completely removed.  Scale bar 250\,nm.}%
\label{fig:fib-sem-wave}%
\end{figure*}

Serial section microscopy has been used for over a century to reconstruct volumetric anatomy of biological samples \citep{born_plattenmodellirmethode_1883}.  Beyond its classical application in biology, zoology, and medical research, serial sectioning in combination with electron microscopy (EM) has become the standard method to reconstruct dense neural connectivity of animal nervous systems at synaptic resolution \citep{briggman_volume_2012, plaza_connectomics_2014, lichtman_connectomics_2014}.  A resolution of less than 10\,nm per pixel is necessary to separate individual neural processes and to recognize chemical synapses.
Sample preparation and data acquisition at this resolution are highly sensitive procedures and, as a result, imaging noise and artifacts during acquisition can be minimized at best, but not entirely avoided.
In this paper, we will focus on two major acquisition modalities for large 3D electron microscopy that are used in EM connectomics: high throughput serial section transmission EM \citep[ssTEM;][]{bock_temca_2011} and block face scanning EM with focused ion beam milling \citep[FIB-SEM;][]{xu_closer_2011,knott_fib_2008,heymann_fib_2006}.  While we have developed our methods with a strong focus on these two modalities, we expect them to generalize well to other applications.

\subsection{Serial Section Transmission Electron Microscopy}
A series of ultra-thin sections is generated by cutting the plastic embedded specimen using an ultra-microtome with a diamond knife.  Section ribbons are collected on tape
\citep{hayworth_atum_2006} or manually.  Manual collection in particular bears the the risk of ordering mistakes that in practice occur frequently.    The nominal section thickness ranges between \UnitPair{30}{90}{nm} which defines the axial resolution.  Yet, discontinuous operation of the ultra-microtome and precision limits of the instrument cause variations of section thickness between and within sections.  Shearing forces applied by the knife and during collection introduce deformations to individual sections.  Section folds, tears, and staining artifacts further complicate the comparison of sections in the series and require that sections be aligned after imaging \citep{saalfeld_elastic_2012}.  However, compared with block face SEM as discussed in the next paragraph, ssTEM has two major advantages: (1) sections can be post-stained which results in improved contrast of structures of interest (\eg{} synapse T-bars), and (2) imaging is performed in transmission mode which enables high acquisition speed and significantly higher in-plane resolution at high signal to noise ratio.

\subsection{Focused Ion Beam Scanning Electron Microscopy}
Block face scanning EM follows a cycle of imaging the block face of a plastic-embedded specimen with a
scanning electron microscope~(SEM) followed by material removal until complete acquisition of the
specimen. In FIB-SEM, focused ion beam~(FIB) milling is used for material removal. This procedure, in practice, generates inhomogeneous $z$-spacing and non-planar block faces leading to
distorted volumes \citep{boergens_controlling_2013,jones_investigation_2014}. These distortions
exhibit a wave-like evolution of height variances throughout the acquired data and can be severe
enough to seriously impede the correct reconstruction of small neural processes (\cf{}
\cref{fig:fib-sem-wave}). FIB-SEM has two major advantages over the previously discussed ssTEM\@:
(1) focused ion beam milling enables significantly higher axial resolution than physical sectioning, which enables the acquisition of isotropic volumes at less than $(\text{10\,nm})^\text{3}$ voxel size, and (2) fully automatic integration of serial imaging and milling in the vacuum chamber of the microscope bears a lower risk for variances in image quality and provides better initial section alignment and correct section order.

\subsection{Contribution}
Extending our previous work~\citep{hanslovsky_post-acquisition_2015}, we developed methods for the identification and correction of ordering mistakes as well as planar and non-planar axial distortions through image-based signal analysis without the need for any further apparatus or physical measurements~(\cref{sec:method}). We  thoroughly assess efficacy and efficiency in virtual ground truth experiments and demonstrate their applicability to real world problems~(\cref{sec:experiments}). We publish all our methods as open source libraries for the ImageJ distribution Fiji and for deployment in a high performance parallel computing environment~\citep{zaharia_spark:_2010}.

\section{Related Work}
\label{sec:related-work}
While, to the best of our knowledge, post-acquisition order correction for serial section microscopy
has not yet been addressed in a rigorous way, several methods exist for measuring or correcting section thickness or
spacing.
\citet{de_groot_comparison_1988} reviews four different methods for estimating section thickness,
all of which require additional physical measurements, specialized apparatus, or even
destructive modifications of previously acquired sections, rendering the proposed methods
impractical or even impossible for certain imaging modalities, \eg{} block face SEM\@. Similarly,
\citet{jones_investigation_2014} introduce an artifact as a fiducial mark from which section
thickness can be estimated in FIB-SEM acquisitions under the assumption of planar
sections. \citet{berlanga_three-dimensional_2011} correct small volumes by evening out top and
bottom surfaces that have been manually annotated by the user and transforming the whole series
accordingly by a single transformation, which fails to capture varying
thickness. \citet{boergens_controlling_2013} reduce non-planar distortions during acquisition by
using measurements of the intensity of the ion beam to control the FIB-SEM milling process. In
addition, they estimate section thickness post-acquisition by adjusting $z$ coordinates such that
the peaks of auto-correlations in several $xz$-cross-sections have the same half-width in both
dimensions. \citet{sporring_estimating_2014} assume strictly isotropic data with thickness
distortions and compare pairwise similarity measures of adjacent sections with a reference curve
that is estimated from in-section pixel intensities. Planar spacing between pairs of adjacent
sections is then determined by evaluating the inverse of the reference curve at the measured
similarities. Global consistency cannot be guaranteed as only pairs of adjacent sections are considered
for estimating spacing. Furthermore, varying image or tissue properties cannot be captured due to
the estimation of the reference from cross-sections perpendicular to the distortion axis.

Our methods are unique among previous approaches in that they do not require any additional
measurements, apparatus, or user annotations, and allow for the correction of section order as well
as both planar and non-planar axial distortions. Contrary to other solely image based methods, our
axial distortion estimates are not based on pairs of sections only, thus improving global
consistency across the image series. Furthermore, varying imaging or tissue properties can be
captured by local estimates of the reference signal along the axis of distortion.

\section{Method}
\label{sec:method}

In the following, we describe our image-based methods for the correction of continuous and discontinuous non-planar axial distortions in serial section microscopy. Our only assumptions are---true for correct section order and spacing---monotonic decrease of pairwise similarity of sections with distance and local constancy of the shape of the similarity function~(\cref{sec:similarity-measure}). Violations of these assumptions indicate wrong section order or spacing. We will describe in detail how coordinate space is transformed to re-establish correctness of these assumptions, and thereby correcting section order mistakes~(\cref{sec:section-order-correction,sec:z-spacing}), planar $z$-spacing~(\cref{sec:z-spacing}), and non-planar $z$-spacing~(\cref{sec:non-planar}).

\subsection{Similarity Measure}
\label{sec:similarity-measure}

We define pairwise similarity $s(P_{z_1},P_{z_2})$ of section $P_{z_1},P_{z_2}\in{} I$, indexed by their respective positions $z_1,z_2$ along the $z$-axis within an image series that has correct section order and spacing, as a symmetric function that decreases monotonically with distance $|z_1-z_2|$:
\begin{align}
  s: & \;I\times{}I \rightarrow [0,1] \subset \mathbb{R} \label{eq:similarity-definition} \\
     & \;(P_{z_1},P_{z_2}) \rightarrow s(P_{z_1},P_{z_2}) = f(|z_1-z_2|) \\
     & \;|z_1-z_2| < |z_3-z_4| \implies f(|z_1-z_2|) > f(|z_3-z_4|) \label{eq:similarity-monotonicity} \\
     & \;s(P_{z_1},P_{z_2}) = s(P_{z_2},P_{z_1}) \label{eq:similarity-symmetry}
\end{align}
For a series of $Z$ sections, all pairwise similarities are stored in a $Z\times{}Z$ matrix
denoted by $\rhom$ such that
\begin{align}
  \rhom(z_1,z_2) = s(P_{z_1},P_{z_2}).
\end{align}
By definition, $\rhom{}$ is a symmetric matrix. In practice, we use noisy surrogate measures for the inaccessible ideal $s$ such that \cref{eq:similarity-monotonicity} may not hold for long distances. Thus, for deformation
estimation, we ignore measurements for which $|z_1-z_2| > r$, for a user specified $r$ that depends on the data set.

We implemented three similarity measures: (1) the Pearson product-moment correlation coefficient (PMCC) for aligned series, (2) the best block matching coefficient (BBMC) for approximately aligned series, and (3) the percentage of true positive feature matches under a transformation model (inlier ratio) for unaligned data.
In our experiments~(\cref{sec:experiments}), we used PMCC and feature inlier ratio.
\subsubsection{Pearson Product-Moment Correlation Coefficient}
\label{sec:pmcc}
The PMCC of two statistical samples $A, B: |A| = |B| = N$ is defined as
\begin{align}
  \label{eq:pmcc}
  \rho_{AB}(A,B) = \frac{\text{cov}(A,B)}{\sqrt{\text{var}(A)\text{var}(B)}} \in [-1,1]
\end{align}
with sample co-variance
\begin{align}
  \label{eq:cov}
  \text{cov}(A,B) = \frac{1}{N}\sum_{i,j}\left(A_{ij}-\mu_A\right)\left(B_{ij}-\mu_B\right),
\end{align}
where $\mu_A=\frac{1}{N}\sum_{i,j}A_{ij}$ is the sample mean, and \text{var}(A) = \text{cov}(A,A) is the sample variance.  PMCC is invariant to changes of the mean and variance of samples $A$ and $B$ and therefore robust against
contrast and gain variations across the image series. In order to comply with
\cref{eq:similarity-definition}, we use
\begin{align}
  \label{eq:pmcc-non-negative}
  s(A,B) = \tilde{\rho}_{AB} = \max(\rho_{AB},0) \in [0,1].
\end{align}

\subsubsection{Best Block Matching Coefficient}
\label{sec:block-matching}
Similarity estimates using PMCC require the series to be perfectly aligned which, in practice, is not always guaranteed.  We therefore implemented an alternative similarity measure that is robust against small local translations, the average over local best block matching coefficients~(BBMC).  For any rectangular region $R_{z_1} \subset P_{z_1}$, the best correspondence $R^{*}_{z_2} \subset P_{z_2}$ is determined by maximizing pairwise PMCC over a set of correspondence candidates $R_{z_2} \subset P_{z_2}$ of the same width and height, sampled in a small radius around the center of the region. The pairwise similarity of sections $P_{z_1}$ and $P_{z_2}$,
\begin{align}
  \label{eq:similarity-calculation-offset}
  \rhom(z_1,z_2) = \frac{1}{N}\sum_{R_{z_1}} \; \max_{R_{z_2}} \;\;s(R_{z_1},R_{z_2}),
\end{align}
is the average of all pairwise similarities between $R_{z_1}$ and corresponding $R^{*}_{z_2}$, where $N$ is the total number of regions $R_{z_1}$ within $P_{z_1}$.

\subsubsection{Inlier Ratio}
\label{sec:inlier-ratio}
Even BBMC requires that the series is approximately aligned.  In ssTEM series, however, approximate alignment is often not available, and aligning the series may be impossible because
the correct order of sections has not yet been established.  To recover the correct order of
sections, we need a similarity measure that is independent of alignment.  Using transformation
invariant features, we match interest points across pairs of sections.  We then use a variant of RANSAC in combination with a least squares local trimming estimator \citep{fischler_random_1981,saalfeld_estimator_2009} to estimate a
model $M$ that transforms one set of interest points onto the other.  The estimator groups all matches into inliers $\mathcal{I}$ that conform with $M$ and outliers $\mathcal{O}$ that do not~($\mathcal{I}\cap\mathcal{O}=\varnothing$).  The similarity of two sections is then given by the inlier ratio
\begin{align}
  \label{eq:inlier-ratio}
  s(\cdot) = \frac{|\mathcal{I}|}{|\mathcal{I}\cup\mathcal{O}|} \in [0,1].
\end{align}
For our experiments, we use SIFT~\citep{lowe_distinctive_2004}.
Where interest point detection and matching are part of the image alignment pipeline, \citep[\eg][]{saalfeld_elastic_2012}, this similarity can be extracted at virtually no cost.

\subsection{Section Order Correction}
\label{sec:section-order-correction}
\begin{figure}
	\includegraphics[width=\linewidth]{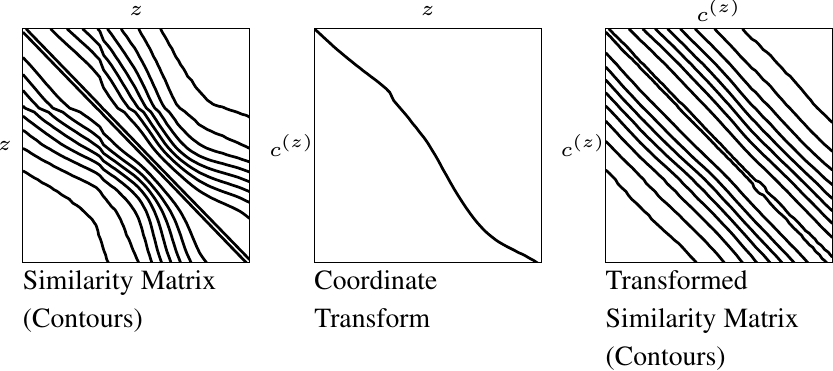}
    \caption{Warp coordinate space such that contour-lines of similarity matrix
        $\rhom(c(z_1),c(z_2))$ are parallel to diagonal.}\label{fig:warp-space}
\end{figure}
Incorrect section order breaks the monotonicity assumption for similarity measures. With pairwise similarity as a proxy for distance between sections, visiting every section in the correct order is equivalent to visiting every section exactly once on the shortest path possible based on distances derived from pairwise similarity. This can be formulated as an augmented traveling salesman problem~\citep[TSP;][]{voigt_handlungsreisende_1831,applegate_traveling_2011}. To that end, we represent the image sections $\{P_i|i=1,\hdots,Z\}$ as vertices $\mathcal{V}=\{1,\hdots,Z\}$ of a fully connected graph $\mathcal{G}=(\mathcal{V},\mathcal{E})$ with edges $\mathcal{E}=\mathcal{V}\times\mathcal{V}$. The weight $w((e_1,e_2))=w((e_2,e_1))\ge 0$ associated with each edge $e = (e_1,e_2)\in \mathcal{E}$ represents the non-negative, symmetric distance between two vertices $e_1$ and $e_2$~(sections). The derived distance is a monotonically increasing function of pairwise similarity $s$: 
\begin{align}
  \label{eq:similarity-to-distance}
  w(e) =&\; f(\rhom(e_1,e_2)) \\
  f :&\; [0,1] \rightarrow \mathbb{R}_{\ge 0} \nonumber \\
     &\; s \rightarrow f(s) \nonumber \\
     &\; \tilde{s} < s \implies f(\tilde{s}) > f(s) \nonumber %
\end{align}
With the addition of a ``start'' vertex $\tilde{\mathcal{V}}=\{0\}$ and zero distance edges
$\tilde{\mathcal{E}}=\{(0,i),(i,0) : \forall i \in \mathcal{V}\}$,
$w(\tilde{e}) = 0~\forall\tilde{e} \in \tilde{\mathcal{E}}$, establishing correct section order is equivalent to solving the TSP for the augmented graph
\begin{align}
  \label{eq:tsp-augmented-graph}
  \tilde{\mathcal{G}} = (\mathcal{V}\cup\tilde{\mathcal{V}},\mathcal{E}\cup\tilde{\mathcal{E}}).
\end{align}

\subsection{Simultaneous Section Spacing and Order Correction}
\label{sec:z-spacing}%
\newcommand\deltaz{\Delta{}z}%

We observed that TSP-sorted series occasionally contain small mistakes such as flipped section pairs.
Pairwise comparison alone does not capture global consistency if the similarity measure is too noisy to
reliably distinguish between pairs of sections~(\cref{fig:result-section-sort-align}) because similarity to any neighbor higher than
first order is completely ignored. We therefore developed a method that, assuming that sections are in approximately correct order, compares shapes of complete similarity matrices to
determine globally consistent order of sections and their relative spacing.

Based on the assumption of monotonically decreasing pairwise similarity and local constancy of the
similarity decay, we formulate an optimization problem that simultaneously
estimates a coordinate transformation $c(z)$ that maps from original section index to real valued section position, a
scalar factor $m(z)$ to compensate for the influence of uncorrelated noise in individual sections to their pairwise similarity scores, and the
``true'' similarity $\similarityEstimate(\cdot)$. Correct section order can be
established by sorting $c(\cdot)$ in increasing order. We summarize all variables,
parameters and measurements in \cref{tab:variables}.

\begin{table}
    \caption{Description of variables and parameters introduced in \crefrange{eq:sse-fit}{eq:optimization}}
    \begin{tabularx}{\linewidth}{l|X}
      Input & \\
      \hline
      $\rhom(z_1,z_2)$ & Symmetric matrix containing measures of similarity for all pairs of sections indexed by $z_1$ and $z_2$. \\
      \multicolumn{2}{l}{}\\
      Variable & \\
      \hline
      $z$,$\zref$,$i$ & Indices referencing \mbox{(sub-)sections} within the data. \\
      $\zmap$ & Coordinate transformation mapping from original coordinate index to corrected
                  coordinate space. \\
      $\zmult$ & Quality assessment for each section to adjust for noise.\\
      $\rhomTransformed(\,\cdot\,)$ & $\rhom(\,\cdot\,)$, corrected by m and warped by
                                      c:\\
      &\mbox{$\rhomTransformed\left(\zmap[z_1],\zmap[z_2]\right)=\zmult[z_1]\times \zmult[z_2]\times\rhom(z_1,z_2)$} \\
      $\similarityEstimate_{\zref}(\deltaz)$ &  Estimate of the similarity curve based on estimates in a local neighborhood
                            around $\zref{}$, sampled at integer coordinates, evaluated at $\deltaz
                                               \in \mathbb{R} $. \\
      \multicolumn{2}{l}{}\\
      Parameter & \\
      \hline
      $\fitWindowWeight$ & Neighborhood around $\zref{}$ for estimation of $\similarityEstimate_{\zref}(\deltaz)$. \\
      $\RRangedWindow{z_1,z_2}$ & Range based windowing function to exclude noisy similarity
                                  measures of distant sections.\\
    \end{tabularx}\label{tab:variables}%
\end{table}

We forgo any assumptions other than monotonicity and constancy of shape in a local
neighborhood. Instead, we estimate the similarity $\similarityEstimate_{\zref}(\Delta{}z)$ as a
function of the distance $\Delta{}z$ between two sections for each local neighborhood located at
section $\zref$ defined by $w_f(\zref,z)$~(\cref{eq:sse-fit}). These local estimates capture changes
in tissue and image properties along the $z$-axis. For all $z$ within this neighborhood, the
measured similarities $\rhomTransformed(c(z),c(z)+\deltaz)$ evaluated at integer distances $\deltaz$
from the position of the section $c(z)$ contribute to the similarity function estimate weighted by
$w_f(\cdot)$.  Simultaneously, we warp the coordinate space such that all measured similarities
agree with the function estimate~(\cref{eq:sse-shift}). In terms of the pairwise similarity matrix
that means aligning the contour lines such that they are parallel to the diagonal~(\cf{}
\cref{fig:warp-space}).

Noise in individual sections decreases the pairwise similarity with all other sections in conflict
to what $\similarityEstimate(\cdot)$ suggests and would thus distort the estimate of $c(\cdot)$. Therefore, we
assess the quality $\zmult$ of each section to distinguish between displacement and other noise that
could distort position correction~(\cref{eq:sse-assess}). Using $m(\cdot)$ to lift the according
similarities closer to $\similarityEstimate(\cdot)$ will diminish this effect. On the other hand,
sections that need displacement will not have a consistent bias towards decreased or increased
similarities and remain unaffected by this quality assessment.

The windowing function $\RRangedWindow{\cdot}$ restricts the evaluation of pairwise
similarities to a range~$r$ to avoid estimation based on distant sections whose
similarity measures tend to be unreliable. In general, we define this window using the Heaviside
step function parameterized by range $r$,
\begin{align}
  \RRangedWindow{z_1,z_2}=\theta(r-|z_2-z_1|).
\end{align}
Each of \cref{eq:sse-fit,eq:sse-assess,eq:sse-shift} contribute to a joint objective~(\cref{eq:optimization}) that is optimized over the function estimate $\similarityEstimate(\cdot)$, the quality measure $m(\cdot)$, and the coordinate mapping $c(\cdot)$:%

\begingroup
\addtolength{\jot}{0.25cm}
\begin{align}
  \label{eq:sse-fit}\text{SSE}_{\text{fit}} =& \sum_{\zref}\sum_{z}\fitWindowWeight \\\nonumber
  &\hspace{-1cm}\times\sum_i\RRangedWindow{z,i} \left(\similarityEstimate_{\zref}(i-z) - \rhomTransformed(\zmap ,\zmap +i-z)\right)^2 \\
  \label{eq:sse-shift}\text{SSE}_{\text{shift}} =& \sum_{z}\sum_{\zref} \RRangedWindow{\zref,z} \\\nonumber
                              &\hspace{-1cm}\times\left( \similarityEstimate_{\zref}^{-1} \left( \zmult[\zref] \zmult \rhom(\zref, z) \right) - \left(\zmap - \zmap[\zref]\right)\right)^2\\
  \label{eq:sse-assess}\text{SSE}_{\text{assess}} =& \sum_{\zref}\sum_{z}\RRangedWindow{\zref,z} \\\nonumber
  &\hspace{-1cm}\times\left(\zmult[\zref]\zmult \rhom(\zref, z) - \similarityEstimate_{\zref}(\zmap -\zmap[\zref])\right)^2\phantom{\sum_z} \\ %
  \label{eq:optimization}\similarityEstimate^*,m^*,c^* =& \argmin_{\similarityEstimate,m,c} \text{SSE}_{\text{fit}} + \text{SSE}_{\text{shift}} + \text{SSE}_{\text{assess}}.
\end{align}
\endgroup

We find a local optimum for \cref{eq:optimization} by alternating least squares.  In the benign case that the series is in approximately correct order and that similarity measures capture sensible information about relative distances between sections, this local optimum is typically the correct solution.  We avoid trivial solutions by meaningful regularization:  $m(\cdot)$ tends towards 1, and $c(\cdot)$ is limited by locking the first and last $z$-positions.
If section order is guaranteed to be correct (as in FIB-SEM), then we do not
allow reordering and enforce $c(z+1) - c(z) > 0$ at any iteration. In addition, we
enforce monotonicity of $\similarityEstimate$ during estimation of both $\similarityEstimate$ and
$c$.  More precisely, if any measurement of similarity $\rhom$ between a point located at $c(z)$ and
reference located at $c(\zref)$ violates the monotonicity assumption, this measurement and all
subsequent measurements located at $\bar{z}$ with \mbox{$|c(\bar{z})-c(\zref)| > |c({z})-c(\zref)|$}
are ignored for this iteration.

\subsection{Non-Planar Axial Distortion Correction}
\label{sec:non-planar}

Section spacing estimation~(\cref{sec:z-spacing}) does not require to consider complete sections but can be applied to any sub-volume defined by a local neighborhood in $x$ and $y$ if similarity can be estimated for pairs of sections in that sub-volume.
Hence, non-planar deformation fields can be estimated by solving \cref{eq:optimization} for a grid of independent similarity matrices, each extracted from a local field of view.  If grid locations were optimized independently, local smoothness could not be guaranteed which is particularly objectionable as similarity measures typically degrade with a smaller field of view and become increasingly susceptible to noise.  Coupling terms between the optimization problems at each grid location would enforce local smoothness, but results in a single large optimization problem instead of many independent optimizations.  We therefore implemented a hierarchical approach, starting at a large field of view---the complete section---and increasing grid resolution and locality with every subsequent stage.  At each stage, we enforce local smoothness and suppress the effects of noise at small fields of view with a regularization term
\begin{align}
  \text{SSE}_{\text{reg}} = \sum_z{\Big( c(z) - \big(\lambda c_0(z)+(1-\lambda)\bar{c}(z)\big)\Big)}^2
\end{align}
towards the inferred coordinates $c_0(\cdot)$ at the previous stage.  The impact of regularization is
controlled by parameter $\lambda\in[0,1]$ with $\bar{c}$ being the result of \cref{eq:sse-shift} at
each iteration of the alternating least squares solution of \cref{eq:optimization}.  All optimization
problems at one stage of the hierarchy depend solely on the results of the previous stage which makes it straightforward to parallelize the solution over all grid cells.
The final resolution of the grid, the field of view considered at each stage, the regularization parameters, and the range of interest for pairwise similarity measurement in the $z$-series are exposed as adjustable parameters to the user.

\begin{table*}%
    \centering
    \caption{Summary of section sort experiments. Data were scaled along $xy$ as stated in
        ``Scale'' column. RT is short for run time. The deviations are the absolute
        values of the differences between predicted and ground truth order. The error rate is the
        ratio of sections that were mapped to the wrong position.}\label{tab:section-sort-experiment}
    \csvloop{
        file=table,
        head to column names,
        tabular={l|rrrrrrrrr},
        table head=\textbf{Name} & \textbf{Sections} & \textbf{Image Size} & \textbf{Scale} &
        \textbf{Similarity} & \textbf{RT Similarity} & \textbf{RT TSP} & \textbf{Error Rate} \\\hline,
        separator=comma,
        command=\Name & \Sections & \SizeX\,\texttimes\,\SizeY & \Scale & \SimilarityCalculationMethod &
        \SimilarityCalculationTimeMs\,ms & \TSPOptimizationTimeMs\,ms & \errorRate
    }
\end{table*}
\section{Experiments}%
\label{sec:experiments}%
\newcommand\SSTEMA{\mbox{ssTEM-a}}%
\newcommand\SSTEMB{\mbox{ssTEM-b}}%
\newcommand\FIBSEMA{\mbox{FIB-SEM-a}}%
\newcommand\FIBSEMB{\mbox{FIB-SEM-b}}%

Following the outline of \cref{sec:method}, we first evaluate section order correction, both using
TSP and section spacing estimation~(\cref{sec:exp:section-order-correction}), before we elaborate on
experiments on spacing correction for planar and non-planar
distortions~(\cref{sec:exp:spacing-correction}).

\subsection{Section Order Correction}
\label{sec:exp:section-order-correction}

\begin{figure}[h]
    \centering
    \includegraphics[width=\linewidth]{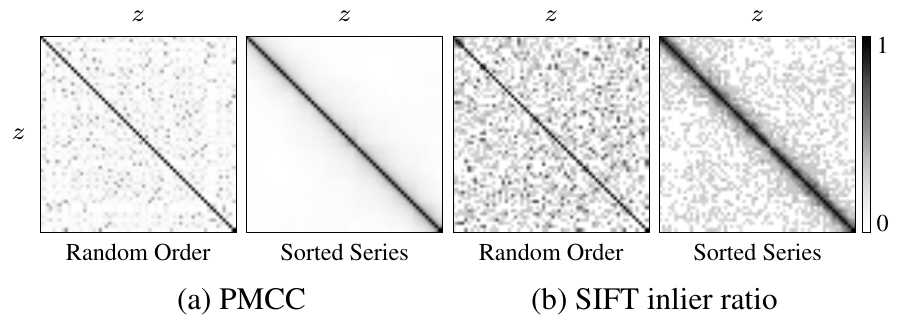}
    \caption{Similarity matrices for randomly permuted \SSTEMA{} series before and after section
        order correction. Similarities were calculated using PMCC~(a) and SIFT
        inlier ratio~(b). 
    }\label{fig:result-section-sort-tem-ncc}
\end{figure}

\begin{figure*}[t]
    \centering
    \newcommand\MyColorMap{\MyGlobalColorMap}
    \includegraphics[width=\linewidth]{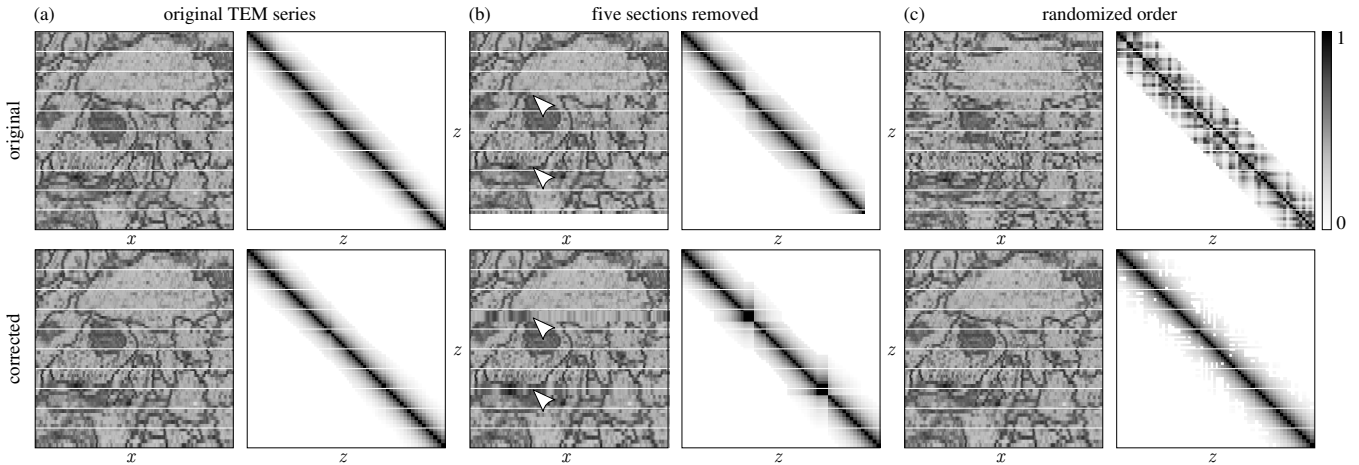}
    \caption{$z$-position correction experiments for \SSTEMA{}: original series~(left), missing
        sections~(center), and randomized order~(right) with a shared coordinate frame in $z$,
        as indicated by the white grid. Top/bottom show an $xz$-cross-section~(left
        sub-column) and corresponding intensity-encoded pairwise similarity matrices~(right
        sub-column) before/after $z$-position correction. Arrows in the center column highlight
        removed sections.}\label{fig:exp-ds3}
\end{figure*}

We begin the evaluation of section order correction with a proof of concept on a small serial
section TEM data set~(\SSTEMA{}\footnote{\label{fn:SSTEMA}ssTEM of \emph{Drosophila melanogaster} CNS, courtesy of
    D. Bock, R. Fetter, K. Khairy, E. Perlman, C. Robinson, Z. Zheng, HHMI Janelia}) of dimensions
\PixelResolution{2580}{3244}{63} and nominal voxel size \PhysicalResolution{4}{4}{40}. We
perturb the correctly ordered series using (1) a completely random permutation and (2) a
permutation that randomly reassings the position of sections within a range of \textpm{}\,4 for the
approaches introduced in \cref{sec:section-order-correction,sec:z-spacing}, respectively. For (1),
we evaluate both PMCC and SIFT inlier ratio as similarity measure.  For (2), we use PMCC only. Similarity
matrices before and after section order correction are shown in
\cref{fig:result-section-sort-tem-ncc,fig:exp-ds3} for TSP (PMCC and SIFT inlier ratio) and section
spacing (PMCC and $xz$-cross-section), respectively. Note that, for (2), section order and $z$-spacing are 
estimated simultaneously and therefore the sorted matrix appears warped. As indicated in \cref{tab:section-sort-experiment}, TSP re-established the correct section order.  The run times for optimization of the TSP problems are negligible compared to the
time required to extract pairwise similarities. Shorter run times for solving the TSP in the PMCC
experiment indicate that, with PMCC, the problem is easier due to better
similarity measures.  PMCC is superior to SIFT inlier ratio as a similarity measure for well aligned
series.  Even for larger examples, the run time for the TSP solution remains short, \eg{}
3\,ms for 2051 sections~(data not shown).  All experiments were carried out on a Dell Precision T7610
workstation using the TSP solver \emph{concorde}~\citep{applegate2006concorde}.

\begin{figure}[h]%
    \includegraphics[width=\linewidth]{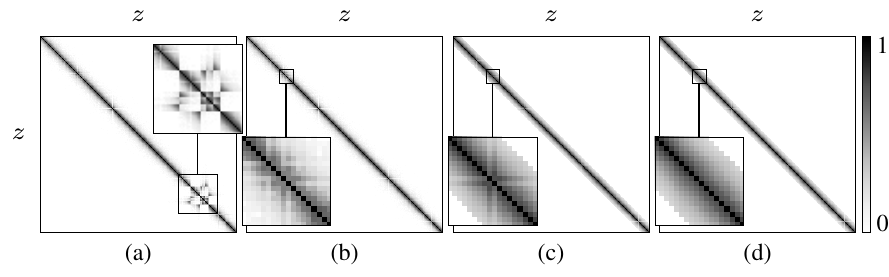}%
    \caption{Section order correction for \SSTEMB{}. Inlier ratio
        matrix for original sequence~(a) and after
        correction~(b). The major disturbance (bottom right)
        could be resolved but two sections remain flipped~(magnified view). This becomes more
        apparent in the PMCC matrix of the aligned
        series~(c). Repeated TSP correction resolves this
        remaining issue~(d).}
    \label{fig:result-section-sort-align}
\end{figure}

With this successful proof of concept at hand, we proceed with section order correction of a longer
section series~(\SSTEMB{}\footnoteref{fn:SSTEMA}).  We chose an unaligned series of 251 complete sections for which we manually curated the correct section order.  The objective of the experiment was to re-establish correct section order from an initially unaligned series with ordering mistakes.
We therefore extracted the SIFT inlier ratio matrix from the unaligned series and estimated section order via TSP.  The solution included small pairwise ordering mistakes.  However, these disturbances were sufficiently local to enable elastic alignment~\citep{saalfeld_elastic_2012} of the corrected series and to extract a PMCC similarity matrix.  We then used the TSP method to estimate order from the PMCC similarities~(\cref{fig:result-section-sort-align}), decreasing the number of misplaced sections from 2 (0.80\%

\subsection{Spacing Correction}
\label{sec:exp:spacing-correction}

Similar to the experiments for section order correction, we start with a proof of concept, followed by an extensive
experiment for the evaluation of non-planar distortion correction using an artificial ground truth
deformation on a real world data set.

\subsubsection{Section Spacing Correction}
\label{sec:exp:section-spacing-correction}

For the evaluation of section spacing correction, we correct and visually inspect distortions within
two data sets: \SSTEMA{} and \FIBSEMA{}\footnote{\label{fn:FIBSEMA}FIB-SEM of \emph{Drosophila melanogaster} CNS, courtesy of K. Hayworth, H.Hess, C. Shan Xu, HHMI Janelia \citep{hayworth_fib_2015}} with dimensions
\PixelResolution{2048}{128}{1000} and nominal voxel size \PhysicalResolution{8}{8}{2}. The
latter is an excerpt of a larger data set with dimensions chosen such that axial distortions can be considered approximately planar.

\Cref{fig:exp-ds3} shows $xz$-cross-sections and the according similarity matrices before and after
section spacing correction for (a)~the original \SSTEMA{} data set, (b) sections
20, 21, 22, 46, 48 removed, and (c) randomized section order. Our experiments
show that for the original data set, $z$-spacing varies between \UnitPair{0.6}{1.6}{px}
(\UnitPair{24}{64}{nm}). Section spacing correction of (b) and (c) is evaluated by comparing
the estimated transformations with the result of (a) as ``ground truth''. The estimated transformation for
(b) correctly stretches the data where sections were removed and deviates (absolute value)
from the ground truth by 0.13\,px (5.2\,nm) on average, and not more than 0.28\,px
(11.2\,nm). Sections removed for this experiment do not contribute to the evaluation. For the
simultaneous order and spacing correction~(c), we measure an absolute deviation from the ground
truth of 0.044\,px (1.76\,nm) on average, and not more than 0.13\,px (5.2\,nm). All
\SSTEMA{} section spacing correction experiments finished in 0.6\,s (similarity matrix
calculation) and 0.4\,s (inference, 100 iterations) on a Dell Precision T7610 workstation.

\begin{figure}[h]
    \newcommand\MyColorMap{\MyGlobalColorMap}
    \includegraphics[width=\linewidth]{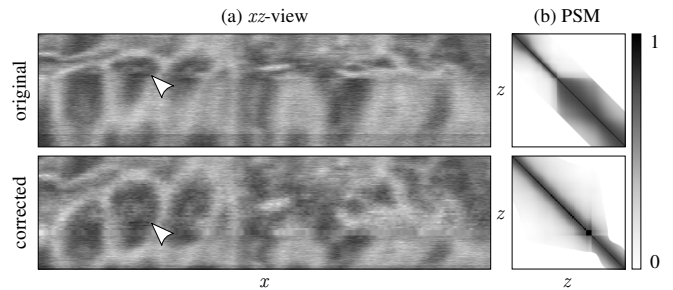}
    \caption{$z$-position correction experiment for \FIBSEMA{}: Top/bottom show an
        $xz$-cross-section~(a) and corresponding intensity-encoded pairwise similarity
        matrices~(PSM;b) before/after $z$-position correction. (a) and (b) share the same coordinate
        frame in $z$. Arrows highlight areas that are visually stretched or compressed in the
        original acquisition and appear biologically plausible after
        correction.}\label{fig:exp-fibsem-1}
\end{figure}

We observe stronger distortions in \FIBSEMA{} as shown in \cref{fig:exp-fibsem-1}~(top). Stretched/condensed regions are highlighted in an $xz$-cross-section and appear in the respective
similarity matrix as regions with slow/fast decay of similarity. After section spacing
correction~(\cref{fig:exp-fibsem-1} bottom), the corrected $xz$-cross-section appears homogeneously sampled and
similarity decay is approximately constant.  The estimated section spacing
varies between 0.14\,px and 10.2\,px, or 0.28\,nm and 20.4\,nm. On the Dell
Precision T7610 workstation used for this experiment, similarity matrix estimation and
inference~(150 iterations) took 62.3\,s and 49.4\,s, respectively.

\subsubsection{Non-Planar Distortion Correction}
\label{sec:exp:non-planar}

\begin{figure*}
    \centering
    \includegraphics[width=\linewidth]{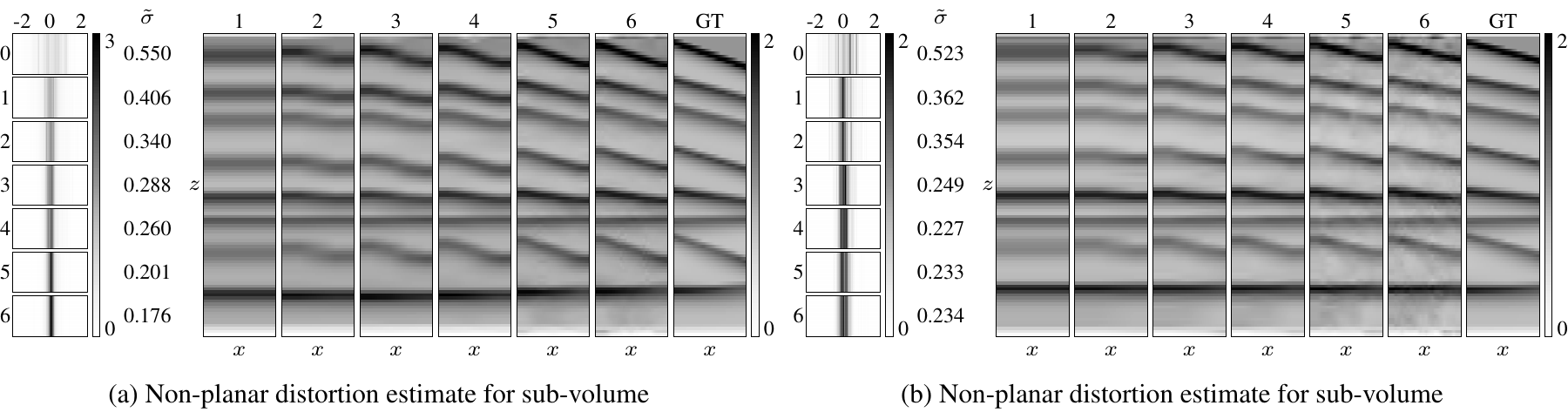}
    \caption{Normalized histograms~(left) of differences between estimated transformation and ground
        truth~(GT) and visualization of the estimated transformation for all stages and GT\@ for
        experiments on both sub-volumes (a) and (b) via $xz$ cross-sections of the
        gradient~(right). Histogram bins range from \UnitRange{-2}{2}{px}, with maximum
        counts of 3 and 2 for (a) and
        (b), respectively. The gradients range from \UnitRange{0}{2}{px}.}%
    \label{fig:non-planar-ground-truth-experiment}
\end{figure*}

We evaluated the performance of non-planar deformation correction against synthetic ground truth.  To that end, we applied synthetic non-planar axial distortion to a distortion free reference series, estimated the distortion with our method (\cref{sec:non-planar}), and compared the estimate with the synthetic ground truth.  Since distortion free volumes do not exist, we had to first correct the original image volume using the same non-planar axial distortion correction method.  The resulting series, from the perspective of our method, is free of distortions.  To compensate for the apparent bias in this approach, we run our experiment not only in the original orientation but permute the coordinate axes such that the new axial dimension falls into the unprocessed image plane.

The data used in this experiment is a subset of \FIBSEMB\footnoteref{fn:FIBSEMA}
with dimensions
\PixelResolution{4000}{2500}{2100} and voxel resolution \PhysicalResolution{8}{8}{2}.  Initial non-planar axial distortion correction was
distributed onto 60 compute nodes with 16 cores each and took 120 minutes to finish.
We scaled the corrected series along the $z$-axis by a factor of 0.25 resulting in an isotropic volume of
\PixelResolution{4000}{2500}{525} from which we extracted two sub-volumes: (a) 100 complete $xy$-sections starting at $z$\,=\,25, and (b) 100 $xz$-cross-sections
of dimension 3000\,\texttimes\,475\,px$^\text{2}$ starting at $y$\,=\,1000.  For (b), we flipped the $y$- and $z$-axes
such that the synthetic distortion can be consistently applied along the $z$-axis.  Our synthetic distortion model is this: Randomly oriented planes superimposed with trigonometric functions act as
attractors that shift the coordinates towards the attractor along the $z$-axis as a monotonically decreasing function of the
distance to the attractor along $z$. This generates waves and plateaus that approximately resemble phenomena that we observed in the original volume before pre-correction. We then applied non-planar distortion correction to the synthetically deformed series as described in \cref{sec:non-planar}. We compare estimated and
ground truth distortions at every stage of the hierarchical solution and show histograms of the pixel-wise differences
~(\cref{fig:non-planar-ground-truth-experiment}).
Since we are not interested in low frequency distortion of the volume, we map each estimate onto the ground truth using a linear transformation that minimizes the squared difference of corresponding look-up table entries within local support defined by a Gaussian window
with $\sigma=(\sigma_x,\sigma_y,\sigma_z)$.

\newcommand\MyNonPlanarGroundTruthExperiment{09}
The evolution of the estimated distortion for each of the sub-volumes is shown shown in
\cref{fig:non-planar-ground-truth-experiment}. For intuitive visualization, the gradient is
displayed. Starting at a complete field of view and a resolution of 1\,px$^\text{2}$ in $x$
and $y$ at stage~1~(planar estimate), the field of view/resolution is decreased/increased by a factor
of two in both $x$ and $y$ with every
sub-sequential stage which allows for a more accurate estimate of the deformation. At the same time,
noise in the data will have a stronger influence on smaller fields of view~(\cf{}
\cref{fig:non-planar-ground-truth-experiment}, stage 6) and sets a limit to the resolution at
which the deformation can be estimated. The histograms of differences did not improve 
after (a) stage~6 or (b) stage~4.
We chose $\sigma$\,=\,$(\infty,\infty,\text{120\,px})$ for the Gaussian window to estimate the
linear transformation. The mean of differences between estimate and ground truth is approximately
zero for all stages.  We therefore used the standard deviation of the error $\tilde\sigma_i$ for
stage~$i$ including the baseline $i=\text{0}$ as a quality indicator.  Smaller $\tilde\sigma_i$ means better estimates of the ground truth.  For (a), we found $\tilde\sigma_\text{0}$\,=\,0.550\,px and $\tilde\sigma_\text{6}$\,=\,0.176\,px, and for (b), we found $\tilde\sigma_\text{0}$\,=\,0.523\,px and $\tilde\sigma_\text{4}$\,=\,0.227\,px.  As expected, non-planar axial distortion correction considerably decreased the distortion of the series in both experiments.

\section{Discussion}
\label{sec:discussion}

We developed novel methods to address two previously unsolved problems: (1) establish the correct order of unordered section series, (2) compensate for planar and non-planar axial distortion.  We demonstrated through extensive experiments that our methods work reliably and with high
accuracy and efficiency on both ssTEM and FIB-SEM data.  We went beyond pure proof of
concept and showed that our methods are applicable to and perform well on large real world data sets.

In large ssTEM series, the combination of automatic alignment and series sorting has the potential to greatly reduce the need for manual intervention.  Non-planar axial distortion correction addresses the peculiar wave-problem in FIB-SEM which, we believe, will have a strong impact on the future application of FIB-SEM for high resolution 3D reconstruction.

In this work, we made only mild assumptions about the data, \ie monotonic decrease of pairwise similarity and local constancy of the shape of the similarity curve.  While this means that our methods can be applied
to a wide range of data, we predict that many problems would benefit from
domain specific modeling.  For example, explicit modeling of FIB-SEM-waves has the potential to further increase
the accuracy of the estimated deformation field.  We will work on these ideas in our future research.

\section{Acknowledgements}
\label{sec:acknowledgements}

This work was supported by HHMI.
We thank Davi Bock,  Ken Hayworth, Harald Hess,
Rick Fetter, Khaled Khairy, Eric Perlman, Camenzind Robinson, Shan Xu, and Zhihao Zheng for data and valuable discussion.

\bibliographystyle{natbib}
\bibliography{main}

\end{document}